\pdfoutput=1
\documentclass[11pt]{article}[final]

\usepackage{acl}

\usepackage{times}
\usepackage{latexsym}

\usepackage[T1]{fontenc}
\usepackage[utf8]{inputenc}

\usepackage{microtype}

\usepackage{inconsolata}

\usepackage{graphicx}
\usepackage{array}
\usepackage{amsmath}
\usepackage{bm}
\usepackage{algorithm}
\usepackage{algpseudocode}
\usepackage{amsmath}
\usepackage{multirow}
\usepackage{indentfirst}
\usepackage{subfigure}
\usepackage{epsfig}
\usepackage{epstopdf}
\usepackage{graphicx}
\usepackage{url}
\usepackage{xspace}
\usepackage{booktabs}
\usepackage{subeqnarray}
\usepackage{subcaption}
\usepackage[most]{tcolorbox}
\usepackage{color}
\usepackage{amssymb}

\newcommand{\paratitle}[1]{\vspace{1.5ex}\noindent\textbf{#1}}

\newcommand{\eg}{\emph{e.g.,}\xspace}

\newcommand{\ignore}[1]{}

\title{Reinforced Informativeness Optimization for Long-Form Retrieval-Augmented Generation}

\author{
    \textbf{Yuhao Wang}\textsuperscript{\rm 1,3}\thanks{~~The work was done during the internship at Baidu.}\quad
    \textbf{Ruiyang Ren}\textsuperscript{\rm 1,3}\thanks{~~Corresponding authors.}\quad
    \textbf{Yucheng Wang}\textsuperscript{\rm 2}\quad
    \textbf{Wayne Xin Zhao}\textsuperscript{\rm 1,3}\footnotemark[2]\\
    \textbf{Jing Liu}\textsuperscript{\rm 2}\footnotemark[2] \quad
    \textbf{Hua Wu}\textsuperscript{\rm 2}\quad 
    \textbf{Haifeng Wang}\textsuperscript{\rm 2}
    \\
    \textsuperscript{1}Gaoling School of Artificial Intelligence, Renmin University of China\\
    \textsuperscript{2}Baidu Inc. \\
    \textsuperscript{3}Beijing Key Laboratory of Research on Large Models and Intelligent Governance \\
    \{yh.wang500, reyon\_ren\}@outlook.com, batmanfly@gmail.com
}
\begin{document}
\maketitle
\begin{abstract}
Long-form question answering (LFQA) requires open-ended long-form responses that synthesize coherent, factually grounded content from multi-source evidence. This makes reinforcement learning (RL) reward design critical. The reward must be verifiable for faithful grounding and stable optimization. However, many standard rewards assume a unique target with an exact-match notion of correctness, which fits short-form QA and math but breaks in LFQA. As a result, current RAG systems still lack verifiable reward mechanisms, yielding unstable feedback signals and suboptimal optimization outcomes.
We propose RioRAG, a framework for reinforced verifiable informativeness optimization.
First, it defines informativeness as a measurable and externally verifiable objective for RL.
Second, RioRAG uses nugget-centric verification with cross-source checks to enable self-evolution of smaller LLMs and to provide denser, action-discriminative rewards that mitigate reward sparsity and stabilize optimization.
This formulation avoids handcrafted supervision for the policy model and strong teacher-model distillation, relying instead on externally verifiable feedback.
Experiments on LongFact and RAGChecker show that RioRAG achieves higher factual recall and faithfulness, establishing verifiable reward modeling as a foundation for trustworthy long-form RAG.
Our codes are available at \url{https://github.com/RUCAIBox/RioRAG}.
\end{abstract}

\section{Introduction}

Long-form question answering (LFQA) represents a crucial step toward enabling AI systems to deliver comprehensive and factually reliable responses by generating elaborate and multi-sentence answers, conditioning language models on input queries~\cite{stelmakh2022asqa}.
Retrieval-augmented generation~(RAG) has emerged as a compelling paradigm for such knowledge-intensive tasks, as it grounds generation in factual content retrieved from external corpora~\cite{ren2023investigating,wang2024rear}.
However, producing reliable long-form answers remains challenging since large language models (LLMs) should synthesize information from multiple retrieved sources into coherent and factual paragraphs~\cite{zhao2023survey}.

\begin{figure}
\centering
\includegraphics[width=0.44\textwidth]{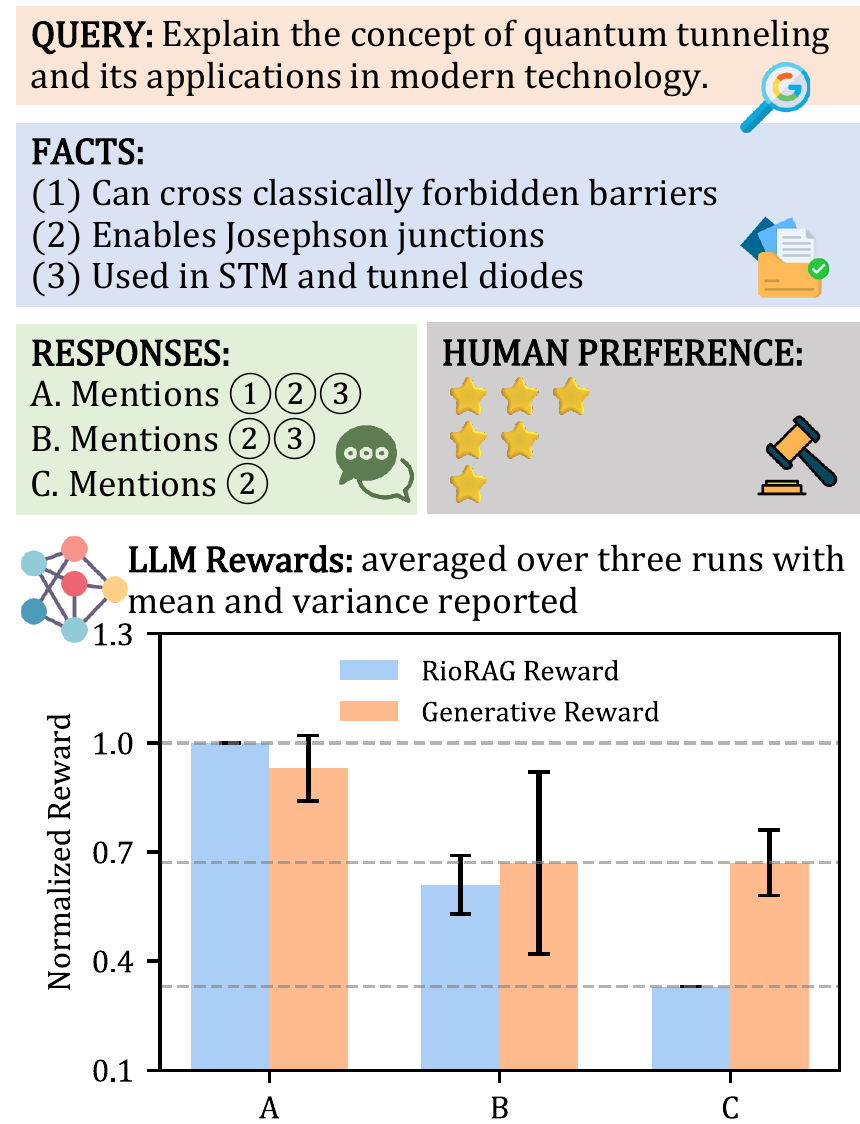}
    \caption{Existing long-form reward modeling exhibits (i) instability, with high variance when evaluating the same response multiple times, and (ii) unreliability, as their score rankings often differ from human judgments across responses.}
    \label{fig:case}
\end{figure}

While RAG mitigates hallucination in long-form generation by grounding models on retrieved evidence~\cite{ren2025self, li2025search,wang2025unveiling}, several challenges remain unresolved.
First, LLMs often misuse factual information~\cite{gao2023enabling, wang2026bee}, either due to residual hallucination or failure to extract the correct evidence from lengthy retrieved documents.
Besides, generated answers frequently exhibit limited informativeness~\cite{ren2025llm}, failing to comprehensively incorporate and reuse the available evidence.
Recent efforts such as prompt engineering and template-based supervised fine-tuning (SFT) partially alleviate these issues~\cite{DBLP:conf/acl/00040LMZHLZ25}, yet they rely heavily on strong annotation signals and external guidance.
Moreover, these methods restrict models’ self-evolution and often reduce generalization and diversity~\cite{shao2024deepseekmath}, making them less adaptable to LFQA tasks across diverse domains.

Recently, reinforcement learning (RL) has improved LLMs in math and short-form QA by optimizing outcome-based feedback~\cite{DBLP:journals/corr/abs-2510-08964,DBLP:journals/corr/abs-2512-21625,DBLP:conf/aaai/ZhanTWLWS26}.
However, RL for LFQA is bottlenecked by reward design.
Many standard rewards assume an exact-match notion of correctness, which does not apply to open-ended long-form answers.
Rule-based outcome reward modelings~(ORMs) are difficult to specify for diverse responses~\cite{guo2025deepseek,li2025search}.
Moreover, long-form scoring is often unstable and hard to verify, even under well-guided evaluation protocols~\cite{zhang2024longreward}.
Reward models can also miss key evidence under long references, yielding misleading signals (Figure~\ref{fig:case}).
Such instability can effectively sparsify the reward signal, further suppressing self-exploration and self-evolution during RL~\cite{guo2025deepseek}.

To address these challenges, we propose RioRAG, a \underline{R}einforced \underline{I}nformativeness \underline{O}ptimization-based RL method for long-form RAG.
RioRAG aims to improve factual coverage with stable, verifiable reward learning.
First, it defines informativeness as an externally verifiable objective and derives rewards from evidence support rather than heuristic lexical rules.
Second, it employs a nugget-centric hierarchical verifier with length-adaptive scoring to provide fine-grained and action-discriminative feedback.
This design mitigates reward sparsity and enables self-evolution of smaller policy models, without relying on handcrafted policy supervision or strong teacher-model distillation (\eg sequential SFT).
Extensive experiments on two published benchmarks, LongFact and RAGChecker, with zero-shot evaluation show that the RioRAG achieves superior performance compared with a series of state-of-the-art methods, demonstrating the effectiveness of the proposed innovations.

\section{Task Formulation}

In this work, we define \emph{informativeness} not as response length or stylistic richness, but as {information coverage under the RAG setting}: how completely an answer covers the distilled, evidence-grounded key points supported by retrieved sources. We refer to these fine-grained factual units as \emph{nuggets}, following prior nugget-based grounded evaluation~\cite{lajewska2025ginger}. Concretely, a nugget is a concise, reusable unit of information, such as a short fact, claim, or QA-style statement, that can be explicitly verified against source content and used to assess whether key information is covered.

LFQA extends conventional SFQA to generate coherent, factual, and detailed multi-paragraph responses (\eg explanations or reports).
Given a user query $q$ and a large web corpus $\mathcal{D} = \{d_1, d_2, \dots, d_N\}$, 
a retriever $R$ retrieves a subset of relevant documents $\mathcal{D}_q \subseteq \mathcal{D}$, 
and a generator $G$ first produces a complete response sequence $y = G(q, \mathcal{D}_q)$ conditioned on both the query and the retrieved evidence.
The response $y$ naturally contains a reasoning part and a final answer part, which we separate by a predefined delimiter:
\[
[\, r_{1:T} \;\Vert\; a_{1:M} \,] = y,
\]

RioRAG follows an ORM formulation. 
Both evaluation and reward computation are based solely on the final answer $a$, 
while the reasoning content in $y$ (\eg chain-of-thought tokens) is ignored by the reward model.
Formally, the objective of long-form RAG is to generate an answer $a$ that maximizes factual correctness, informativeness, and coherence with respect to $\mathcal{D}_q$.
In RioRAG, we approach this objective from an RL perspective, 
where the model learns to maximize a verifiable informativeness reward derived from the generated outcome.

\section{Method}
\label{sec:method}

\begin{figure*}
    \centering
    \includegraphics[width=0.99\linewidth]{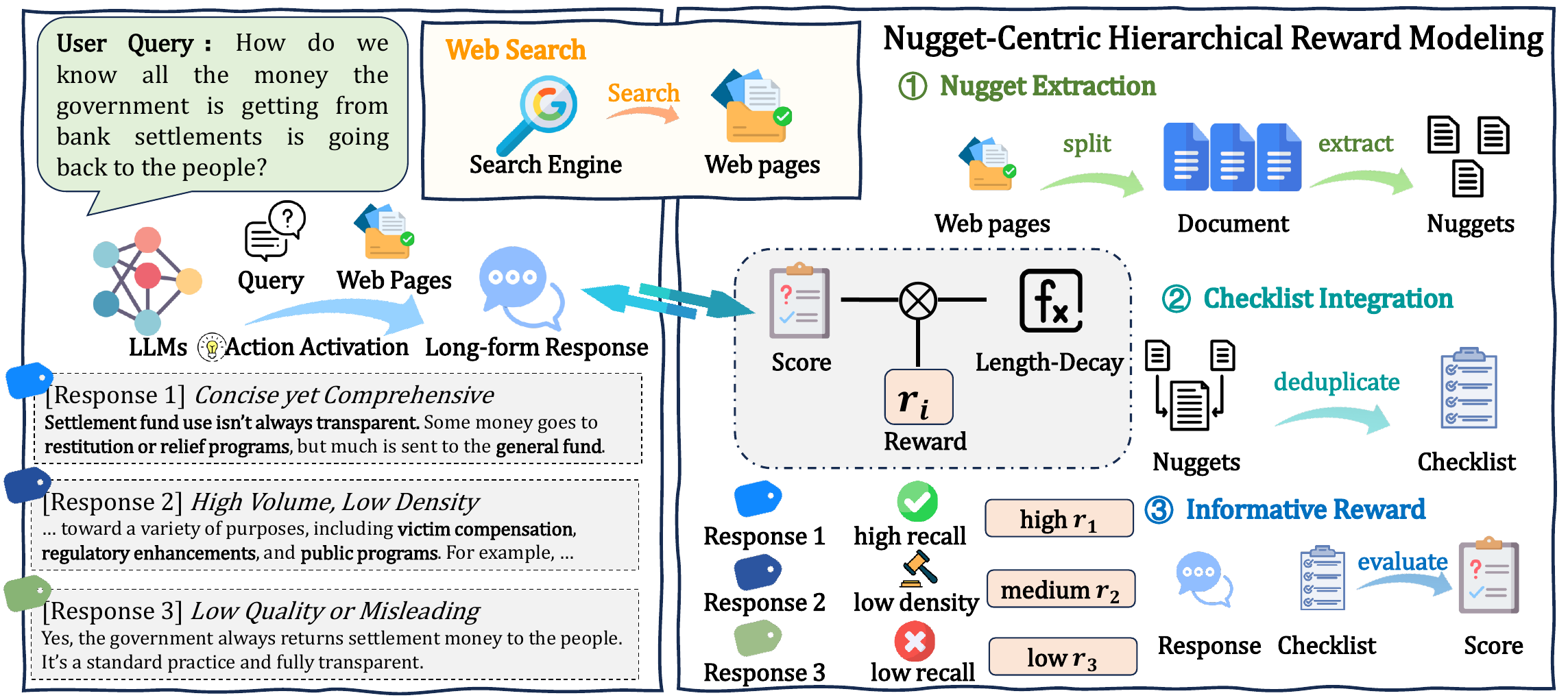}
    \caption{Overall illustration of the proposed RL-based RioRAG framework.}
    \label{fig:framework}
    \vspace{-2ex}
\end{figure*}

% \subsection{RioRAG Overview}

Figure~\ref{fig:framework} shows the overall pipeline of RioRAG. 
Given a query, RioRAG retrieves diverse and up-to-date web documents and generates long-form answers through reinforcement learning. 
The framework consists of two main components: 
(i) \textit{reinforced informativeness optimization}, which maximizes factual coverage reward, and 
(ii) \textit{nugget-centric hierarchical reward modeling}, which provides fine-grained, verifiable feedback based on evidence-level nuggets. 
Together, these components enable stable and unsupervised optimization of long-form RAG systems.

\subsection{Reinforced Informativeness Optimization}

Previous works mainly relied on SFT, which depends on pre-defined templates or powerful teacher models and thus limits generalization and adaptability to open-domain queries~\cite{menick2022teaching}.  
Inspired by the recent success of RL in enhancing LLMs through self-exploration in tasks like math and coding~\cite{yang2025qwen3,DBLP:conf/aaai/LiZZFW26,liu2026drugtrail}, we extend this paradigm to the LFQA setting of RAG.  
RioRAG aims to enable models to self-improve by interacting with real-time web knowledge and optimizing generation quality without external supervision.

\paratitle{Training Data Construction.}
Existing RAG systems often depend on labeled data from strong teacher models and static retrieval corpora, both of which limit generalization to new domains and fail to reflect real-time web knowledge.  
To overcome these limitations, we reconstruct the training setup using ELI5~\cite{fan2019eli5}, retaining only queries while removing human-written answers.  
For each query, we dynamically retrieve the top-$K$ webpages to form $\mathcal{D}_q$, ensuring that the model learns from current and diverse online sources.  
This design allows the model to explore and adapt to evolving web information under RL optimization, improving factuality and robustness in long-form generation.

\paratitle{Stable Reinforcement Optimization.}
Conventional RL optimization for long-form text often suffers from unstable gradients and inconsistent reward scaling, 
especially when model outputs vary significantly in length or quality.
Inspired by relative normalization strategies that stabilize optimization in long-horizon reasoning tasks~\cite{guo2025deepseek}, 
we adopt Group-wise Relative Policy Optimization (GRPO)~\cite{shao2024deepseekmath} to improve training stability.  
Given $G$ sampled completions $\{o_1, \dots, o_G\}$ with corresponding rewards $\{r_1, \dots, r_G\}$, 
GRPO normalizes rewards within each sampled group:
\[
\begin{aligned}
A_i &= \frac{r_i - \mu_r}{\sigma_r + \epsilon}, \\
\mu_r &= \tfrac{1}{G}\!\sum_{j=1}^{G} r_j,\quad
\sigma_r = \sqrt{\tfrac{1}{G}\!\sum_{j=1}^{G}(r_j - \mu_r)^2}.
\end{aligned}
\]
This group-wise normalization mitigates reward variance across samples, 
stabilizes policy updates, and leads to smoother reward improvement during long-form RL training.

\paratitle{Action Activation.}
Previous approaches introduce new delimiter tokens (\eg \texttt{<think>}) to structure reasoning, which hinders warm-start from instruction-tuned LLMs and biases learning toward format imitation.  
We observe that even base models naturally support Markdown formatting.  
Hence, we leverage Markdown headers to separate reasoning and answers without adding new tokens.  
This lightweight design preserves pretrained priors, improves optimization stability, and provides a structured foundation for our subsequent hierarchical reward modeling.

\subsection{Nugget-based Verifiable Reward Modeling}

Existing ORMs for long-form generation are often non-verifiable and unstable~\cite{ru2024ragchecker}, as they rely on heuristic or model-based scoring that may hallucinate or fluctuate across runs.  
To overcome these challenges, we propose a \textit{nugget-based hierarchical reward modeling} approach that decomposes reward computation into three verifiable stages.

\paratitle{(1) Factual Nugget Extraction.}
For each retrieved document $d_i \in \mathcal{D}_q$, we identify \emph{factual nuggets} as concise evidence statements. Each nugget $n_{ij}$ is extracted as a short clause from $d_i$, forming a nugget set:
\[
\mathcal{N}(\mathcal{D}_q) = \bigcup_{d_i \in \mathcal{D}_q} \text{Extract}(d_i).
\]

\paratitle{(2) Evidence Checklist Synthesis.}  
The extracted nuggets are clustered and merged into a unified checklist that captures all distinct factual evidence relevant to query $q$.  
This aggregation removes redundancy and enables consistent cross-document evaluation:
\[
\mathcal{C}(q, \mathcal{D}_q) = \text{MergeCluster}(\mathcal{N}(\mathcal{D}_q)).
\]

\paratitle{(3) Informativeness Assessment.}  
Given a generated output $o$, we compute the informativeness reward by measuring the proportion of checklist nuggets covered in $o$:
\[
\mathcal{I}(q, o, \mathcal{D}_q) = 
\frac{|\{ n \in \mathcal{C}(q, \mathcal{D}_q) \mid n \subseteq o \}|}
{|\mathcal{C}(q, \mathcal{D}_q)|}.
\]
Each matched nugget can be explicitly traced to its source document, ensuring transparent and verifiable reward computation.

\paratitle{Length Decay.}  
To prevent overestimation from excessively long responses, we apply a length-decay normalization on the reward computation.  
The decay term is applied only to the final answer segment, leaving the reasoning tokens unaffected to preserve the model’s deliberative process.  
Formally, the length-adjusted reward is defined as:
\begin{equation}
\label{eq:decay}
r_i =
% \mathcal{I}(q, o_i, \mathcal{D}_q) =
\begin{cases}
s_i \exp\!\big[-k ((l - l_0)/\tau)^m \big], & l > l_0,\\[3pt]
s_i, & \text{otherwise,}
\end{cases}
\end{equation}
where $s_i$ is the base informativeness score, $l$ denotes the answer length, 
$l_0$ is the threshold of unpenalized length, $\tau$ controls the decay rate, 
and $k, m$ regulate the sharpness of the decay curve.  
This design maintains factual compactness without penalizing intermediate reasoning, ensuring the reward focuses on the informativeness and precision of final outputs.

This hierarchical design stabilizes reward estimation, mitigates noise from heuristic scoring, and provides a reproducible, fact-grounded supervision signal for reinforcement learning.

\subsection{Discussion}

\begin{table}[t]
\centering
\small
\begin{tabular}{lcc}
\toprule
Setting & Runtime & Complexity \\
\midrule
Generative Reward & 1.19 & $\mathcal{O}((q + nd)^2)$ \\
RioRAG sequential & 2.32 & $\mathcal{O}(n(q + d)^2 + q^2)$ \\
RioRAG w/ parallel & 0.98 & -- \\
RioRAG w/ async & 0.72 & -- \\
\bottomrule
\end{tabular}
\caption{Reward computation cost comparison. Runtime is measured in seconds per training sample. Here $q$ is the query length and $d$ is document length.}
\label{tab:runtime}
\end{table}

\paratitle{Novelty.}
RioRAG’s novelty comes from turning open-ended LFQA evaluation into explicit, verifiable reward computation.
(1) {From vague holistic scoring to checklist credit assignment.} Instead of asking a judge model to output a coarse overall score under ambiguous rules, RioRAG converts informativeness into nugget-aligned checklist items. Each item becomes a concrete scoring point with clear credit.
(2) {From long-context judging to short-context verification.} RioRAG first condenses retrieved evidence into nugget-aligned checklists. It then compares the model output against the checklists and aggregates item-wise scores into the final reward. This avoids long-context degradation, yields denser feedback, and supports self-evolution without strong teacher-model distillation (\eg sequential SFT).

\paratitle{Efficiency.}
Although RioRAG adds checklist extraction and nugget-based evaluation, these steps operate on compact inputs (single documents or short checklists), keeping token counts low.
As a result, RioRAG is comparable to generative reward baselines in the sequential setting, and becomes faster with parallel execution.
Reward computation can also be executed asynchronously and pipelined with RL training, so it does not block gradient updates in practice.

\paratitle{Relation to Recent Reward Design.}
Recent studies have explored rubric- or checklist-based reward design for long-form generation and related RL settings~\cite{gunjal2507rubrics}. RioRAG is complementary to this line of work, but is specifically designed for {long-form RAG}, where reward reliability is challenged by lengthy retrieved contexts and multi-source evidence integration. Our reward construction first extracts factual nuggets from retrieved documents in a decomposed manner and then merges them into a unified checklist, reducing the effective context length for verification and improving reward reliability under long contexts. We also distinguish RioRAG from work that uses retrieval mainly as a verifier during training rather than as the inference-time generation setting~\cite{chen2025learning,chen2025train}. In contrast, our setting focuses on {long-form QA with RAG at inference time}, where the model must synthesize answers from retrieved multi-source evidence during generation.

\section{Experiment}
\label{sec:experiments}
In this section, we detail the experimental setup, present the main results, and further support our findings with ablation studies and in-depth analysis.

\subsection{Experimental Setup}

\subsubsection{Datasets}
We use the ELI5 dataset~\cite{fan2019eli5} as the training source, but only its question corpus is used without reference answers. This avoids overfitting to concise, single-perspective annotations and better reflects real retrieval-augmented settings. A total of 10K questions are randomly sampled for RL training.

For evaluation, we adopt two long-form QA benchmarks: {LongFact}~\cite{weilong} and {RAGChecker}~\cite{ru2024ragchecker}. 
LongFact covers 38 domains consolidated into 8 major categories, with answers annotated by atomic factual units for fine-grained factual verification. 
RAGChecker includes 10 public datasets spanning 4K questions, designed to assess factual grounding and retrieval-based answer quality across multiple dimensions.
Further dataset details are provided in Appendix~\ref{sec:appendix_datasets}.

\subsubsection{Evaluation Metrics}
Following standard LFQA studies~\cite{fan2019eli5}, we use \textit{fact recall} (FR) and \textit{information density} (ID) as the main metrics, measuring factual completeness and conciseness.
We further report RAGChecker’s multi-dimensional metrics~\cite{ru2024ragchecker}, including \textit{faithfulness}, \textit{hallucination}, and \textit{context utilization}, to capture both factual reliability and retrieval effectiveness.
Importantly, these RAGChecker metrics are used only for evaluation and are not provided to the model in any form during training, ensuring an objective assessment.
Full metric definitions are presented in Appendix~\ref{sec:appendix_metrics}.

\begin{table*}[t]
\setlength\tabcolsep{2.7pt}
\centering
\small
\scalebox{0.92}{
\begin{tabular}{lcccccccccccccccccc}
\toprule
\multicolumn{1}{c}{\multirow{2.5}{*}{\textbf{Method}}} & \multicolumn{2}{c}{{Science}} & \multicolumn{2}{c}{{Tech.}} & \multicolumn{2}{c}{{Medicine}} & \multicolumn{2}{c}{{Law}} & \multicolumn{2}{c}{{Culture}} & \multicolumn{2}{c}{{Events}} & \multicolumn{2}{c}{{Commun.}} & \multicolumn{2}{c}{{Lifestyle}} & \multicolumn{2}{c}{\textbf{Average}} \\
\cmidrule(r){2-3} \cmidrule(r){4-5} \cmidrule(r){6-7} \cmidrule(r){8-9} \cmidrule(r){10-11} \cmidrule(r){12-13} \cmidrule(r){14-15} \cmidrule(r){16-17}\cmidrule(r){18-19}
\multicolumn{1}{c}{} & \textbf{FR} & \textbf{ID} & \textbf{FR} & \textbf{ID} & \textbf{FR} & \textbf{ID} & \textbf{FR} & \textbf{ID} & \textbf{FR} & \textbf{ID} & \textbf{FR} & \textbf{ID} & \textbf{FR} & \textbf{ID} & \textbf{FR} & \textbf{ID} & \textbf{FR} & \textbf{ID} \\

\midrule
\multicolumn{19}{c}{\textit{Prompt-based Methods}}\\
% \cmidrule{1-16}
{Direct-RAG}  & 45.3 & 54.2 & 60.6 & 69.3 & 45.7 & 61.1 & 46.9 & 31.4 & 45.4 & 55.4 & 51.2 & 44.3 & 55.5 & 60.2 & 48.7 & 69.8 & 49.6 & 53.5\\
{Chain-of-Thought}  & 55.5 & 51.6 & 53.7 & 71.8 & 46.4 & 58.8 & 45.2 & 30.8 & 47.1 & 61.2 & 48.5 & 46.3 & 54.6 & 59.0 & 54.6 & 70.8 & 50.6 & 54.1 \\
Chain-of-Note   & 45.0 & 52.6 & 58.4 & 71.5 & 43.6 & 58.7 & 42.3 & 27.1 & 40.9 & 56.9 & 47.7 & 46.5 & 47.8 & 56.9 & 48.8 & 71.0 & 46.0 & 52.5\\
{GopherCite}  & 54.4 & 56.6 & 63.8 & 73.9 & 56.2 & 54.1 & 55.6 & 33.7 & 53.3 & 60.6 & 48.9 & 46.4 & 61.5 & 64.0 & 54.1 & 72.5 & 55.9 & 56.1\\
\midrule
\multicolumn{19}{c}{\textit{Supervised Fine-tuning based Methods}}\\
\multicolumn{1}{l}{Chain-of-Note}  & 65.3 & 123.3 & 50.0 & 129.9 & 77.4 & 130.6 & 57.5 & 93.8 & 67.5 & 134.3 & 52.5 & 80.2 & 64.8 & 147.8 & 64.6 & 114.7 & 62.2 & 119.5\\
\multicolumn{1}{l}{GopherCite}   & 59.2 & 121.5 & 58.4 & 145.5 & 74.4 & 118.8 & 59.3 & 80.2 & 69.0 & 146.0 & 68.4 & 101.3 & 61.8 & 144.1 & 60.0 & 103.5 & 63.2 & 119.7  \\
\midrule
\multicolumn{19}{c}{\textit{RL-based Methods}}\\
\multicolumn{1}{l}{DPO} & 59.0 & 106.7 & 66.2 & 137.0 & 60.7 & 96.4 & 65.7 & 61.8 & 69.2 & 115.0 & 56.6 & 83.2 & 60.5 & 122.9 & 61.3 & 113.9 & 62.8 & 102.7  \\
\textbf{RioRAG} & \textbf{69.7} & \textbf{146.7} & \textbf{63.3} & \textbf{170.4} & \textbf{77.4} & \textbf{142.1} & \textbf{77.9} & \textbf{113.4} & \textbf{78.0} & \textbf{120.4} & \textbf{71.6} & \textbf{117.7} & \textbf{75.2} & \textbf{170.7} & \textbf{61.5} & \textbf{144.9} & \textbf{72.8} & \textbf{138.8}  \\

\bottomrule
\end{tabular}}
\caption{The results on eight broader categories of LongFact benchmark with the average results of the eight categories, where FR denotes fact recall and ID denotes information density.}
\label{tab:longfact}
\end{table*}

\begin{table*}[t]
    \centering
    \small 
\scalebox{0.94}{
    \setlength{\tabcolsep}{4pt} 
    \begin{tabular}{lcccccccc}
        \toprule
        \textbf{Method} & \textbf{Fact-Rec$\uparrow$} & \textbf{Info-Den$\uparrow$} & \textbf{Cont-Util$\uparrow$} & \textbf{Rel-NS$\downarrow$} & \textbf{Irrel-NS$\downarrow$}  & \textbf{Hallu.$\downarrow$} & \textbf{Self-Know$\downarrow$} & \textbf{Faith.$\uparrow$}\\
        \midrule
        \multicolumn{9}{c}{\textit{Prompt-based Methods}}\\
        % \cmidrule{1-16}
        Direct-RAG  & 38.3 & 91.6 & 22.6 & 8.2 & 7.5 & 37.0 & 8.1 & 45.3\\
        {Chain-of-Thought} & 50.4 & 146.5 & 24.3 & 4.6 & 4.2 & 30.2 & 9.7 & 48.0  \\
        {Chain-of-Note} & 38.7 & 144.3 & 18.3 & 6.8 & 5.1 & 53.0 & 6.9 & 35.7  \\
        GopherCite & 51.4 & 138.5 & 26.0 & 5.1 & 4.3 & 29.2 & 10.8 & 47.5 \\

        \midrule
        \multicolumn{9}{c}{\textit{Supervised Fine-tuning based Methods}}\\
        \multicolumn{1}{l}{Chain-of-Note} & 54.2 & 190.2 & 22.7 & 4.3 & 3.7 & 22.6 & 7.8 & 30.2  \\
        \multicolumn{1}{l}{GopherCite} & 62.6 & 209.9 & 26.0 & 5.1 & 4.3 & 29.2 & 10.8 & 52.5  \\
        \midrule
        \multicolumn{9}{c}{\textit{RL-based Methods}}\\
        \multicolumn{1}{l}{DPO} & 61.2 & 149.6 & 26.0 & 5.2 & 6.0 & 27.8 & 8.0 & 53.1  \\
        \textbf{RioRAG} & \textbf{66.3} & \textbf{224.6} & \textbf{27.8} & \textbf{4.3} & \textbf{3.6} & \textbf{20.9} & \textbf{5.0} & \textbf{58.2}\\
        \bottomrule
    \end{tabular}
}
    \caption{Average results across ten domains on the RAGChecker benchmark. Fact-Rec refers to fact recall, Info-Den to information density, Cont-Util to context utilization, Rel-NS and Irrel-NS to relevant and irrelevant noise sensitivity, Hallu. to hallucination, Self-Know to self-knowledge, and Faith. to faithfulness.}

    \label{tab:ragchecker}
\end{table*}

\subsubsection{Baselines}
To evaluate the performance of RioRAG, we conduct comprehensive comparisons with various classical and state-of-the-art baseline methods across different categories, ensuring a thorough understanding of the proposed approach. The baselines are categorized into three groups based on their training paradigms: prompt-based unsupervised methods, supervised fine-tuning~(SFT)-based approaches, and RL-based techniques.
For prompt-based methods, we select GopherCite~\cite{menick2022teaching}, chain-of-thought~\cite{wei2022chain} and chain-of-note~\cite{yu2024chain}. Among SFT-based approaches, we employ chain-of-note and GopherCite with the SFT setting. For RL-based methods, we adopt the Direct Preference Optimization~(DPO)~\cite{rafailov2023direct} framework. All baseline implementations are manually reimplemented with rigorous adherence to identical experimental configurations to ensure a fair comparison.
This evaluation protocol guarantees the reliability of performance benchmarking while controlling for potential confounding factors in implementation differences.

\begin{table*}[t]
    \centering
    \small
    \setlength{\tabcolsep}{2.2pt}
    \scalebox{0.92}{
        \begin{tabular}{lcccccccccccccccccc}
        \toprule
        \multicolumn{1}{c}{\multirow{2.5}{*}{\textbf{Method}}} & \multicolumn{2}{c}{{Science}} & \multicolumn{2}{c}{{Tech.}} & \multicolumn{2}{c}{{Medicine}} & \multicolumn{2}{c}{{Law}} & \multicolumn{2}{c}{{Culture}} & \multicolumn{2}{c}{{Events}} & \multicolumn{2}{c}{{Commun.}} & \multicolumn{2}{c}{{Lifestyle}} & \multicolumn{2}{c}{\textbf{Average}} \\
        \cmidrule(r){2-3} \cmidrule(r){4-5} \cmidrule(r){6-7} \cmidrule(r){8-9} \cmidrule(r){10-11} \cmidrule(r){12-13} \cmidrule(r){14-15} \cmidrule(r){16-17}\cmidrule(r){18-19}
        \multicolumn{1}{c}{} & \textbf{FR} & \textbf{ID} & \textbf{FR} & \textbf{ID} & \textbf{FR} & \textbf{ID} & \textbf{FR} & \textbf{ID} & \textbf{FR} & \textbf{ID} & \textbf{FR} & \textbf{ID} & \textbf{FR} & \textbf{ID} & \textbf{FR} & \textbf{ID} & \textbf{FR} & \textbf{ID} \\

        \midrule
        RioRAG  & \textbf{69.7} & \textbf{146.7} & \textbf{63.3} & \textbf{170.4} & \textbf{77.4} & \textbf{142.1} & \textbf{77.9} & \textbf{113.4} & \textbf{78.0} & \textbf{120.4} & \textbf{71.6} & \textbf{117.7} & \textbf{75.2} & \textbf{170.7} & \textbf{61.5} & \textbf{144.9} & \textbf{72.8} & \textbf{138.8} \\
        \midrule
        w/o Info. Optim. & 34.4 & 79.6 & 53.2 & 147.6 & 32.0 & 92.3 & 33.4 & 66.6 & 40.9 & 87.5 & 43.2 & 76.6 & 43.9 & 106.4 & 36.6 & 97.3 & 39.5 & 90.8 \\
        w/o Nugget Reward & 36.0 & 77.6 & 56.6 & 119.3 & 23.8 & 62.5 & 36.0 & 42.4 & 37.2 & 80.7 & 30.7 & 65.4 & 36.8 & 83.2 & 40.1 & 112.7 & 37.0 & 77.3  \\
        w/o Length Decay  & 57.0 & 99.2 & 65.5 & 139.9 & 58.3 & 109.9 & 62.0 & 88.7 & 56.5 & 87.5 & 43.7 & 63.0 & 45.2 & 70.7 & 68.5 & 108.8 & 56.2 & 91.3  \\
        w/ Generative GRPO & 58.7 & 104.3 & 63.3 & 159.2 & 36.6 & 62.7 & 55.7 & 70.9 & 63.1 & 112.7 & 54.3 & 70.3 & 52.5 & 103.2 & 61.7 & 99.3 & 54.7 & 97.2 \\
        w/ Off-Policy RL & 41.6 & 41.4 & 61.7 & 65.5 & 34.1 & 35.0 & 44.8 & 25.1 & 46.8 & 52.2 & 46.4 & 39.8 & 56.2 & 53.1 & 61.8 & 60.2 & 49.3 & 45.5  \\
        \bottomrule
    \end{tabular}}
    \caption{Results of the RioRAG variants on LongFact.}
    \label{tab:ablation}
\end{table*}

\subsection{Main Results}
The results of different methods evaluated on LongFact and RAGChecker are shown in Table~\ref{tab:longfact} and Table~\ref{tab:ragchecker}. It can be observed that:

(1) Our comprehensive evaluation reveals that SFT-based baselines substantially outperform prompt-based approaches, demonstrating the inherent limitations of prompt engineering in handling complex information synthesis tasks. The proposed RioRAG framework establishes a significant improvement across all metrics. This improvement stems from the reinforced informativeness optimization paradigm, which implements a nugget-centric hierarchical reward mechanism to guide LLMs in processing long-context inputs.

(2) Comprehensive evaluation on RAGChecker demonstrates that RioRAG excels in long-form RAG tasks across multiple critical dimensions, including knowledge point coverage, information density, retrieval utilization, hallucination mitigation, and internal knowledge integration. These results underscore the multidimensional efficacy of the proposed approach.

(3) Compared to off-line RL-based methods such as DPO, RioRAG demonstrates superior performance in long-form reasoning tasks. By leveraging an enhanced on-policy GRPO algorithm, RioRAG enables more comprehensive exploration of potential reasoning strategies during generation, thereby optimizing the LLM more effectively through informativeness-driven reward feedback.

\begin{figure*}[t]
    \centering
    \includegraphics[width=0.99\linewidth]{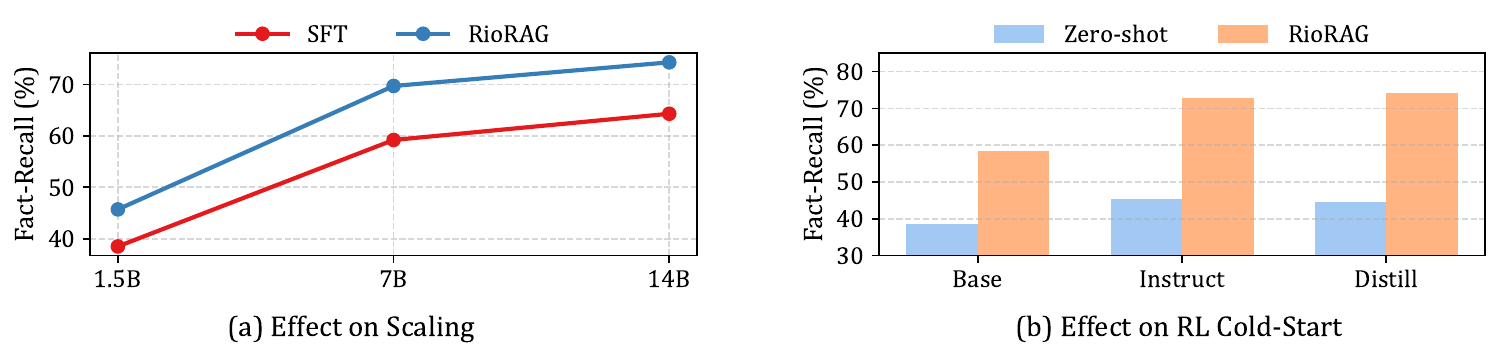}
    \caption{In-depth analysis on scaling law and RL cold-start.}
    \label{fig:scaling}

\end{figure*}
\begin{figure*}
    \centering
    \includegraphics[width=0.99\linewidth]{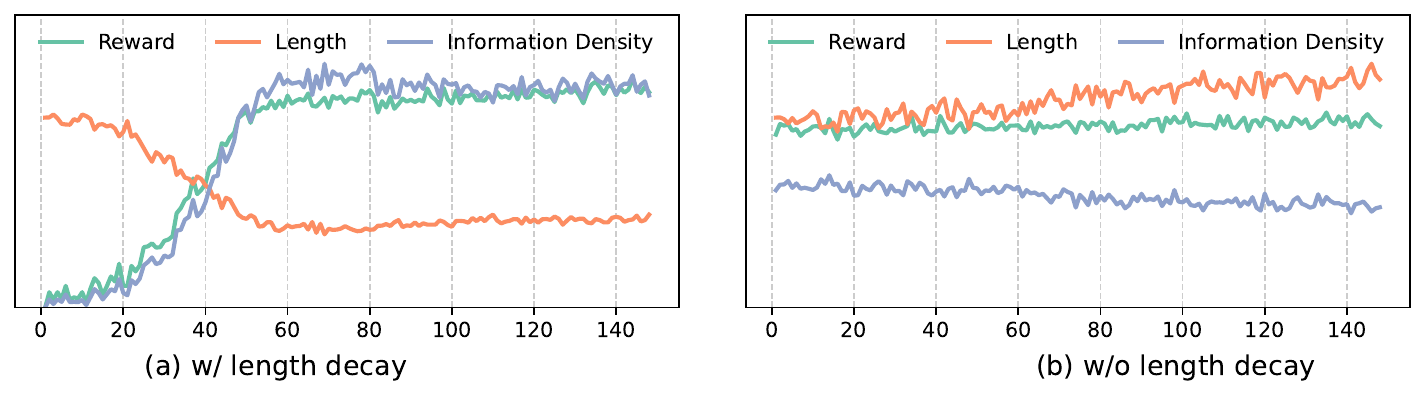}
    \caption{Analysis on co-evolution of generation length and reward during RL training.}
    \label{fig:length}

\end{figure*}

\subsection{Ablation Studies}

In this section, we conduct an ablation study to evaluate the effectiveness of critical strategies in RioRAG comprehensively on LongFact. Here, we consider five variants built on RioRAG for evaluation: (a) \underline{\textit{w/o Info. Optim.}} removes the informativeness-based reward optimization during RL, replaced by direct quality evaluation; (b) \underline{\textit{w/o Nugget Reward}} removes the nugget-wise information extraction and use the full webpage for checklist integration; (c) \underline{\textit{w/o Length Decay}} eliminates the length penalty in Equation~(\ref{eq:decay}); (d) \underline{\textit{w/ Generative GRPO}} denotes the setting where the policy is trained with a 32B generative reward model providing scalar feedback, rather than the verifiable reward used in RioRAG; (e) \underline{\textit{w/ Off-Policy RL}} utilizes an off-policy RL method that employs a static sampling strategy wherein all queries are pre-processed through offline rollouts to generate complete trajectories before being uniformly scored.

Table~\ref{tab:ablation} presents the results for the variants of our method, from which we can observe the following findings: 
(a) The performance drops in \underline{\textit{w/o Info. Optim.}}, demonstrating that using informativeness as the objective for optimization enhances the performance of long-form RAG models through the guidance of reasoning. 
(b) The performance drops in \underline{\textit{w/o Nugget Reward}}, demonstrating incorporating nugget-wise information extraction enables the model to better capture core facts.
(c) The performance drops in \underline{\textit{w/o Length Decay}}, underscoring the critical role of incorporating the length penalty in mitigating excessive response length.
(d) The performance degradation observed in \underline{\textit{w/ Generative RM}} indicates that, without an appropriate pipeline design, unstable and unreliable rewards hinder effective fine-tuning.
(e) The performance significantly drops in \underline{\textit{w/ Off-Policy RL}}, demonstrating that the off-policy method may contain a mismatch between the behavior policy and the target policy.
% compared to the on-policy method GRPO.

\subsection{Further Analysis}

\subsubsection{Scaling Law of RioRAG}
To investigate the scalability characteristics of RioRAG, we conduct a systematic analysis using the Qwen2.5 model with varying parameter sizes~(1.5B, 7B, and 14B). As illustrated in Figure~\ref{fig:scaling} (a), the experimental results demonstrate that RioRAG significantly outperforms SFT at all model scales, with performance consistently improving in accordance with scaling laws.

We can first observe that larger models exhibit improved semantic understanding for both query formulation and webpage relevance assessment. Second, RioRAG benefits from increased model capacity for learning sophisticated retrieval utilization strategies. Third, the enhanced generation capability of larger models enables more effective utilization of retrieved webpages while reducing hallucination risks through better alignment with the reward model's feedback.
Notably, the performance growth curve shows a sublinear relationship between model size and metric improvements, aligning with observations from language model scaling studies~\cite{weiemergent}. This phenomenon suggests that while our RioRAG framework effectively leverages model scale, there exists an upper bound where additional parameters may not proportionally improve RAG performance, which is a critical consideration for practical system deployment.

\subsubsection{Effect on RL Cold-Start}

To examine how model initialization affects RL training, we compare three variants: a base model (Qwen2.5-7B-Base), an instruction-tuned model (Qwen2.5-7B-Instruct), and an R1-distilled model (R1-Distilled-Qwen2.5-7B) incorporating DeepSeek R1’s slow-thinking distillation~\cite{guo2025deepseek}. As shown in Figure~\ref{fig:scaling}(b), the instruction-tuned model gains 24.4\%, and the R1-distilled model achieves the largest improvement of 29.6\% after RL training.

The results show that initialization strongly affects training stability and reward learning. The base model, lacking alignment and reasoning priors, struggles to explore effectively. In contrast, the R1-distilled model benefits from pre-established reasoning patterns, enabling more stable reward estimation and efficient policy updates. This supports that structured reasoning provides a favorable starting point for RL.

\subsubsection{Generation Length and Reward}
To investigate the dynamics of generation length control during RL training, we systematically analyze the interaction between sequence length evolution and reward optimization.
Figure~\ref{fig:length} illustrates the co-evolution of average generation length and reward scores across training steps.

Without length control, the model often produces overly long answers, while the reward score stays almost unchanged. This shows a form of reward hacking, where the model gains higher scores through longer outputs instead of better content. After adding the length-decay term, the model first explores longer responses and then learns to shorten them while keeping rewards stable. The results show that length regularization helps improve both clarity and information density in long-form generation.
\section{Related Work}

\subsection{Long-Form Question Answering}

Research on LFQA has progressed through three shifts: fine-tuned generative models, retrieval-augmented architectures, and human-aligned LLMs. Early abstractive LFQA was enabled by datasets such as ELI5~\cite{fan2019eli5}, showing that seq2seq models can generate plausible long answers with retrieved evidence. More recent systems leverage LLMs with human feedback. Often via RL from human preferences. To produce answers grounded in explicit quotations~\cite{menick2022teaching}. RAG has since become the dominant approach for factual grounding, including training models in web-browsing interfaces~\cite{nakano2021webgpt, qin2023webcpm}. Verifiability can be strengthened by training with explicit source citation~\cite{menick2022teaching}, while post-hoc attribution verifies pre-generated text~\cite{wang2022modeling}. Given LFQA’s open-ended nature, faithfulness to evidence is a central objective~\cite{gao2023enabling, zhao2024tapera}. It can be improved through open-book training with citations~\cite{menick2022teaching} or modeled via probabilistic calibration of answer correctness~\cite{huang2024calibrating}. Complementarily, LLM-based evaluators provide automated quality assessment for LFQA~\cite{han2024rag}.

\subsection{Reinforcement Learning based RAG}

RL has emerged as a promising tool for improving RAG by optimizing retrieval and generation with reward-driven policy updates.
Early work strengthens behavior-cloned web-browsing agents with human-feedback rewards to improve factual alignment~\cite{nakano2021webgpt}.
Subsequent studies design fine-grained rewards for coherence and information gain~\cite{cai2024forag}, or build domain-specific reward models with synthetic supervision to align RAG with human performance~\cite{nguyen2024reward}.
Others use composite reward ensembles to balance answer quality and coverage~\cite{zhang2025rag}.
On the retrieval side, modules can be optimized via group-wise relative policy optimization~\cite{huang2025rag} or multi-agent coordination~\cite{chen2025improving}.
Cost-sensitive retrieval further uses value estimation to decide when to invoke external search under latency--utility trade-offs~\cite{kulkarni2024reinforcement}.
Recent progress, exemplified by DeepSeek-R1~\cite{guo2025deepseek}, has also motivated RL for autonomous retrieval invocation within long reasoning chains~\cite{jin2025search,tang2026towards}.
In contrast, RioRAG introduces nugget-centric hierarchical reward modeling to optimize verifiable informativeness for long-form RAG, without handcrafted policy supervision or strong teacher-model distillation.

\section{Conclusion}
\label{sec:con}

In this work, we address long-form RAG limitations through RioRAG, an RL framework that redefines long-form RAG training via reinforced informativeness optimization with nugget-centric hierarchical reward modeling. RioRAG directly optimizes informativeness through a quantifiable reward design for factual alignment, without the need for scarce training data.
Our experiments on two benchmarks demonstrate that RioRAG fundamentally improves the quality of long-form RAG.
By addressing the core challenges identified in long-form RAG, RioRAG advances the development of trustworthy generative systems for real-world knowledge applications.
Moreover, the success of nugget-level reward modeling suggests that future evaluation frameworks for long-form tasks should prioritize granular factual alignment over surface-level metrics.
Limitations include the current focus on English corpora and reliance on automatic nugget extraction, which may inherit biases from pre-trained models. For future work, we will extend the framework to multilingual settings and investigate human-in-the-loop reward refinement.

\section*{Limitations}
While RioRAG effectively improves the stability and verifiability of long-form RAG training, several limitations remain. First, our current reward primarily targets factual informativeness and does not explicitly capture other aspects of long-form quality, such as linguistic style, coherence, and readability. These factors may further influence human preference alignment and should be considered in future extensions. Second, although we examine models of different scales, computational resources restrict us from scaling beyond 32B parameters. Larger-scale experiments (\eg 72B) could provide deeper insights.

We used AI assistants for minor language polishing; all technical content and results were produced and verified by the authors. Potential risks may arise from imperfect automatic nugget extraction and verification, especially in high-stakes domains.

% Bibliography entries for the entire Anthology, followed by custom entries
%\bibliography{anthology,custom}
% Custom bibliography entries only
\bibliography{custom}

@inproceedings{fan2019eli5,
  title={ELI5: Long Form Question Answering},
  author={Fan, Angela and Jernite, Yacine and Perez, Ethan and Grangier, David and Weston, Jason and Auli, Michael},
  booktitle={Proceedings of the 57th Annual Meeting of the Association for Computational Linguistics},
  pages={3558--3567},
  year={2019}
}

@article{gunjal2507rubrics,
  title={Rubrics as rewards: Reinforcement learning beyond verifiable domains, 2025},
  author={Gunjal, Anisha and Wang, Anthony and Lau, Elaine and Nath, Vaskar and Liu, Bing and Hendryx, Sean},
  journal={URL https://arxiv. org/abs/2507.17746}
}

@inproceedings{lajewska2025ginger,
  title={Ginger: Grounded information nugget-based generation of responses},
  author={{\L}ajewska, Weronika and Balog, Krisztian},
  booktitle={Proceedings of the 48th International ACM SIGIR Conference on Research and Development in Information Retrieval},
  pages={2723--2727},
  year={2025}
}

@article{chen2025learning,
  title={Learning to reason for factuality},
  author={Chen, Xilun and Kulikov, Ilia and Berges, Vincent-Pierre and O{\u{g}}uz, Barlas and Shao, Rulin and Ghosh, Gargi and Weston, Jason and Yih, Wen-tau},
  journal={arXiv preprint arXiv:2508.05618},
  year={2025}
}

@article{chen2025train,
  title={Train for Truth, Keep the Skills: Binary Retrieval-Augmented Reward Mitigates Hallucinations},
  author={Chen, Tong and Asai, Akari and Zettlemoyer, Luke and Hajishirzi, Hannaneh and Brahman, Faeze},
  journal={arXiv preprint arXiv:2510.17733},
  year={2025}
}

@article{shao2024deepseekmath,
  title={Deepseekmath: Pushing the limits of mathematical reasoning in open language models},
  author={Shao, Zhihong and Wang, Peiyi and Zhu, Qihao and Xu, Runxin and Song, Junxiao and Bi, Xiao and Zhang, Haowei and Zhang, Mingchuan and Li, YK and Wu, Y and others},
  journal={arXiv preprint arXiv:2402.03300},
  year={2024}
}

@article{menick2022teaching,
  title={Teaching language models to support answers with verified quotes},
  author={Menick, Jacob and Trebacz, Maja and Mikulik, Vladimir and Aslanides, John and Song, Francis and Chadwick, Martin and Glaese, Mia and Young, Susannah and Campbell-Gillingham, Lucy and Irving, Geoffrey and others},
  journal={arXiv preprint arXiv:2203.11147},
  year={2022}
}

@article{rosenthal2025clapnq,
  title={CLAPnq: C ohesive L ong-form A nswers from P assages in Natural Questions for RAG systems},
  author={Rosenthal, Sara and Sil, Avirup and Florian, Radu and Roukos, Salim},
  journal={Transactions of the Association for Computational Linguistics},
  volume={13},
  pages={53--72},
  year={2025},
  publisher={MIT Press 255 Main Street, 9th Floor, Cambridge, Massachusetts 02142, USA~…}
}

@article{wang2024novelqa,
  title={Novelqa: A benchmark for long-range novel question answering},
  author={Wang, Cunxiang and Ning, Ruoxi and Pan, Boqi and Wu, Tonghui and Guo, Qipeng and Deng, Cheng and Bao, Guangsheng and Wang, Qian and Zhang, Yue},
  journal={arXiv e-prints},
  pages={arXiv--2403},
  year={2024}
}

@article{yang2025qwen3,
  title={Qwen3 technical report},
  author={Yang, An and Li, Anfeng and Yang, Baosong and Zhang, Beichen and Hui, Binyuan and Zheng, Bo and Yu, Bowen and Gao, Chang and Huang, Chengen and Lv, Chenxu and others},
  journal={arXiv preprint arXiv:2505.09388},
  year={2025}
}

@article{zhang2024longreward,
  title={Longreward: Improving long-context large language models with ai feedback},
  author={Zhang, Jiajie and Hou, Zhongni and Lv, Xin and Cao, Shulin and Hou, Zhenyu and Niu, Yilin and Hou, Lei and Dong, Yuxiao and Feng, Ling and Li, Juanzi},
  journal={arXiv preprint arXiv:2410.21252},
  year={2024}
}

@article{li2025search,
  title={Search-o1: Agentic search-enhanced large reasoning models},
  author={Li, Xiaoxi and Dong, Guanting and Jin, Jiajie and Zhang, Yuyao and Zhou, Yujia and Zhu, Yutao and Zhang, Peitian and Dou, Zhicheng},
  journal={arXiv preprint arXiv:2501.05366},
  year={2025}
}

@article{guo2025deepseek,
  title={Deepseek-r1: Incentivizing reasoning capability in llms via reinforcement learning},
  author={Guo, Daya and Yang, Dejian and Zhang, Haowei and Song, Junxiao and Zhang, Ruoyu and Xu, Runxin and Zhu, Qihao and Ma, Shirong and Wang, Peiyi and Bi, Xiao and others},
  journal={arXiv preprint arXiv:2501.12948},
  year={2025}
}

@article{stelmakh2022asqa,
  title={ASQA: Factoid questions meet long-form answers},
  author={Stelmakh, Ivan and Luan, Yi and Dhingra, Bhuwan and Chang, Ming-Wei},
  journal={arXiv preprint arXiv:2204.06092},
  year={2022}
}

@article{xu2024kiwi,
  title={Kiwi: A dataset of knowledge-intensive writing instructions for answering research questions},
  author={Xu, Fangyuan and Lo, Kyle and Soldaini, Luca and Kuehl, Bailey and Choi, Eunsol and Wadden, David},
  journal={arXiv preprint arXiv:2403.03866},
  year={2024}
}

@inproceedings{maia201818,
  title={Www'18 open challenge: financial opinion mining and question answering},
  author={Maia, Macedo and Handschuh, Siegfried and Freitas, Andr{\'e} and Davis, Brian and McDermott, Ross and Zarrouk, Manel and Balahur, Alexandra},
  booktitle={Companion proceedings of the the web conference 2018},
  pages={1941--1942},
  year={2018}
}

@article{nakano2021webgpt,
  title={Webgpt: Browser-assisted question-answering with human feedback},
  author={Nakano, Reiichiro and Hilton, Jacob and Balaji, Suchir and Wu, Jeff and Ouyang, Long and Kim, Christina and Hesse, Christopher and Jain, Shantanu and Kosaraju, Vineet and Saunders, William and others},
  journal={arXiv preprint arXiv:2112.09332},
  year={2021}
}

@inproceedings{qin2023webcpm,
  title={WebCPM: Interactive Web Search for Chinese Long-form Question Answering},
  author={Qin, Yujia and Cai, Zihan and Jin, Dian and Yan, Lan and Liang, Shihao and Zhu, Kunlun and Lin, Yankai and Han, Xu and Ding, Ning and Wang, Huadong and others},
  booktitle={The 61st Annual Meeting Of The Association For Computational Linguistics},
  year={2023}
}

@inproceedings{
liu2026drugtrail,
title={DrugTrail: Explainable Drug Discovery via Structured Reasoning and Druggability\nobreakdash-Tailored Preference Optimization},
author={Yurou Liu and Mingyang Li and Xinyuan Zhu and Rui Jiao and Yiming Dong and Xinyu Tang and Yang Liu and Jieping Ye and Bing Su and Zheng Wang},
booktitle={The Fourteenth International Conference on Learning Representations},
year={2026},
url={https://openreview.net/forum?id=1pAW0y8WLH}
}

@inproceedings{DBLP:conf/nips/ZhanLSG24,
  author       = {Yu{-}Liang Zhan and
                  Zhong{-}Yi Lu and
                  Hao Sun and
                  Ze{-}Feng Gao},
  title        = {Over-parameterized Student Model via Tensor Decomposition Boosted
                  Knowledge Distillation},
  booktitle    = {NeurIPS},
  year         = {2024}
}

@inproceedings{ren2023investigating,
  title={Investigating the factual knowledge boundary of large language models with retrieval augmentation},
  author={Ren, Ruiyang and Wang, Yuhao and Qu, Yingqi and Zhao, Wayne Xin and Liu, Jing and Tian, Hao and Wu, Hua and Wen, Ji-Rong and Wang, Haifeng},
  booktitle={Proceedings of the 31st International Conference on Computational Linguistics (COLING 2025)},
  year={2023}
}

@article{zhao2023survey,
  title={A survey of large language models},
  author={Zhao, Wayne Xin and Zhou, Kun and Li, Junyi and Tang, Tianyi and Wang, Xiaolei and Hou, Yupeng and Min, Yingqian and Zhang, Beichen and Zhang, Junjie and Dong, Zican and others},
  journal={arXiv preprint arXiv:2303.18223},
  volume={1},
  number={2},
  pages={1--124},
  year={2023}
}

@inproceedings{DBLP:conf/aaai/LiZZFW26,
  author       = {Yifan Li and
                  Kun Zhou and
                  Xin Zhao and
                  Lei Fang and
                  Jirong Wen},
  title        = {Analyzing and Mitigating Object Hallucination: {A} Training Bias Perspective},
  booktitle    = {{AAAI}},
  pages        = {6636--6643},
  publisher    = {{AAAI} Press},
  year         = {2026}
}

@inproceedings{DBLP:conf/acl/00040LMZHLZ25,
  author       = {Xinyu Tang and
                  Xiaolei Wang and
                  Zhihao Lv and
                  Yingqian Min and
                  Xin Zhao and
                  Binbin Hu and
                  Ziqi Liu and
                  Zhiqiang Zhang},
  title        = {Unlocking General Long Chain-of-Thought Reasoning Capabilities of
                  Large Language Models via Representation Engineering},
  booktitle    = {{ACL} {(1)}},
  pages        = {6832--6849},
  publisher    = {Association for Computational Linguistics},
  year         = {2025}
}

@inproceedings{ren2025self,
  title={Self-calibrated listwise reranking with large language models},
  author={Ren, Ruiyang and Wang, Yuhao and Zhou, Kun and Zhao, Wayne Xin and Wang, Wenjie and Liu, Jing and Wen, Ji-Rong and Chua, Tat-Seng},
  booktitle={Proceedings of the ACM on Web Conference 2025},
  pages={3692--3701},
  year={2025}
}

@inproceedings{ren2025llm,
  title={LLM-based Search Assistant with Holistically Guided MCTS for Intricate Information Seeking},
  author={Ren, Ruiyang and Wang, Yuhao and Li, Junyi and Jiang, Jinhao and Zhao, Wayne Xin and Wang, Wenjie and Chua, Tat-Seng},
  booktitle={Proceedings of the 48th International ACM SIGIR Conference on Research and Development in Information Retrieval},
  pages={1098--1108},
  year={2025}
}

@inproceedings{DBLP:conf/aaai/ZhanTWLWS26,
  author       = {Yu{-}Liang Zhan and
                  Xinyu Tang and
                  Han Wan and
                  Jian Li and
                  Jirong Wen and
                  Hao Sun},
  title        = {L2V-CoT: Cross-Modal Transfer of Chain-of-Thought Reasoning via Latent
                  Intervention},
  booktitle    = {{AAAI}},
  pages        = {12358--12366},
  publisher    = {{AAAI} Press},
  year         = {2026}
}

@article{li2025viper,
  title={Beyond the Last Frame: Process-aware Evaluation for Generative Video Reasoning},
  author={Li, Yifan and Gu, Yukai and Min, Yingqian and Liu, Zikang and Du, Yifan and Zhou, Kun and Yang, Min and Zhao, Wayne Xin and Qiu, Minghui},
  journal={arXiv preprint arXiv:2512.24952},
  year={2025}
}

@inproceedings{
tang2026towards,
title={Towards High Data Efficiency in Reinforcement Learning with Verifiable Reward},
author={Xinyu Tang and Zhenduo Zhang and Yurou Liu and Xin Zhao and zujie wen and Zhiqiang Zhang and JUN ZHOU},
booktitle={The Fourteenth International Conference on Learning Representations},
year={2026},
url={https://openreview.net/forum?id=sruA4AZmZI}
}

@article{DBLP:journals/corr/abs-2510-08964,
  author       = {Yifan Li and
                  Zhenghao Chen and
                  Ziheng Wu and
                  Kun Zhou and
                  Ruipu Luo and
                  Can Zhang and
                  Zhentao He and
                  Yufei Zhan and
                  Wayne Xin Zhao and
                  Minghui Qiu},
  title        = {Unleashing Perception-Time Scaling to Multimodal Reasoning Models},
  journal      = {CoRR},
  volume       = {abs/2510.08964},
  year         = {2025}
}

@article{DBLP:journals/corr/abs-2512-21625,
  author       = {Xinyu Tang and
                  Yuliang Zhan and
                  Zhixun Li and
                  Wayne Xin Zhao and
                  Zhenduo Zhang and
                  Zujie Wen and
                  Zhiqiang Zhang and
                  Jun Zhou},
  title        = {Rethinking Sample Polarity in Reinforcement Learning with Verifiable
                  Rewards},
  journal      = {CoRR},
  volume       = {abs/2512.21625},
  year         = {2025}
}

@inproceedings{zhao2024tapera,
  title={TaPERA: enhancing faithfulness and interpretability in long-form table QA by content planning and execution-based reasoning},
  author={Zhao, Yilun and Chen, Lyuhao and Cohan, Arman and Zhao, Chen},
  booktitle={Proceedings of the 62nd Annual Meeting of the Association for Computational Linguistics (Volume 1: Long Papers)},
  pages={12824--12840},
  year={2024}
}

@article{kulkarni2024reinforcement,
  title={Reinforcement learning for optimizing rag for domain chatbots},
  author={Kulkarni, Mandar and Tangarajan, Praveen and Kim, Kyung and Trivedi, Anusua},
  journal={arXiv preprint arXiv:2401.06800},
  year={2024}
}

@inproceedings{wang2024rear,
  title={Rear: A relevance-aware retrieval-augmented framework for open-domain question answering},
  author={Wang, Yuhao and Ren, Ruiyang and Li, Junyi and Zhao, Wayne Xin and Liu, Jing and Wen, Ji-Rong},
  booktitle={Proceedings of the 2024 Conference on Empirical Methods in Natural Language Processing},
  pages={5613--5626},
  year={2024}
}

@inproceedings{wang2025unveiling,
  title={Unveiling Knowledge Utilization Mechanisms in LLM-based Retrieval-Augmented Generation},
  author={Wang, Yuhao and Ren, Ruiyang and Wang, Yucheng and Zhao, Wayne Xin and Liu, Jing and Wu, Hua and Wang, Haifeng},
  booktitle={Proceedings of the 48th International ACM SIGIR Conference on Research and Development in Information Retrieval},
  pages={1262--1271},
  year={2025}
}

@inproceedings{wang2026bee,
  title={Bee-rag: Balanced entropy engineering for retrieval-augmented generation},
  author={Wang, Yuhao and Ren, Ruiyang and Wang, Yucheng and Liu, Jing and Zhao, Xin and Wu, Hua and Wang, Haifeng},
  booktitle={Proceedings of the AAAI Conference on Artificial Intelligence},
  volume={40},
  number={40},
  pages={33737--33745},
  year={2026}
}

@article{jin2025search,
  title={Search-r1: Training llms to reason and leverage search engines with reinforcement learning},
  author={Jin, Bowen and Zeng, Hansi and Yue, Zhenrui and Yoon, Jinsung and Arik, Sercan and Wang, Dong and Zamani, Hamed and Han, Jiawei},
  journal={arXiv preprint arXiv:2503.09516},
  year={2025}
}

@article{weiemergent,
  title={Emergent Abilities of Large Language Models},
  author={Wei, Jason and Tay, Yi and Bommasani, Rishi and Raffel, Colin and Zoph, Barret and Borgeaud, Sebastian and Yogatama, Dani and Bosma, Maarten and Zhou, Denny and Metzler, Donald and others},
  journal={Transactions on Machine Learning Research}
}

@article{rafailov2023direct,
  title={Direct preference optimization: Your language model is secretly a reward model},
  author={Rafailov, Rafael and Sharma, Archit and Mitchell, Eric and Manning, Christopher D and Ermon, Stefano and Finn, Chelsea},
  journal={Advances in Neural Information Processing Systems},
  volume={36},
  pages={53728--53741},
  year={2023}
}

@inproceedings{yu2024chain,
  title={Chain-of-Note: Enhancing Robustness in Retrieval-Augmented Language Models},
  author={Yu, Wenhao and Zhang, Hongming and Pan, Xiaoman and Cao, Peixin and Ma, Kaixin and Li, Jian and Wang, Hongwei and Yu, Dong},
  booktitle={Proceedings of the 2024 Conference on Empirical Methods in Natural Language Processing},
  pages={14672--14685},
  year={2024}
}

@article{wei2022chain,
  title={Chain-of-thought prompting elicits reasoning in large language models},
  author={Wei, Jason and Wang, Xuezhi and Schuurmans, Dale and Bosma, Maarten and Xia, Fei and Chi, Ed and Le, Quoc V and Zhou, Denny and others},
  journal={Advances in neural information processing systems},
  volume={35},
  pages={24824--24837},
  year={2022}
}

@article{ru2024ragchecker,
  title={Ragchecker: A fine-grained framework for diagnosing retrieval-augmented generation},
  author={Ru, Dongyu and Qiu, Lin and Hu, Xiangkun and Zhang, Tianhang and Shi, Peng and Chang, Shuaichen and Jiayang, Cheng and Wang, Cunxiang and Sun, Shichao and Li, Huanyu and others},
  journal={Advances in Neural Information Processing Systems},
  volume={37},
  pages={21999--22027},
  year={2024}
}

@inproceedings{weilong,
  title={Long-form factuality in large language models},
  author={Wei, Jerry and Yang, Chengrun and Song, Xinying and Lu, Yifeng and Hu, Nathan Zixia and Huang, Jie and Tran, Dustin and Peng, Daiyi and Liu, Ruibo and Huang, Da and others},
  booktitle={The Thirty-eighth Annual Conference on Neural Information Processing Systems},
  year={2024}
}

@article{nguyen2024reward,
  title={Reward-RAG: Enhancing RAG with Reward Driven Supervision},
  author={Nguyen, Thang and Chin, Peter and Tai, Yu-Wing},
  journal={arXiv preprint arXiv:2410.03780},
  year={2024}
}

@article{chen2025improving,
  title={Improving Retrieval-Augmented Generation through Multi-Agent Reinforcement Learning},
  author={Chen, Yiqun and Yan, Lingyong and Sun, Weiwei and Ma, Xinyu and Zhang, Yi and Wang, Shuaiqiang and Yin, Dawei and Yang, Yiming and Mao, Jiaxin},
  journal={arXiv preprint arXiv:2501.15228},
  year={2025}
}

@article{huang2025rag,
  title={RAG-RL: Advancing Retrieval-Augmented Generation via RL and Curriculum Learning},
  author={Huang, Jerry and Madala, Siddarth and Sidhu, Risham and Niu, Cheng and Hockenmaier, Julia and Zhang, Tong},
  journal={arXiv preprint arXiv:2503.12759},
  year={2025}
}

@inproceedings{cai2024forag,
  title={FoRAG: Factuality-optimized Retrieval Augmented Generation for Web-enhanced Long-form Question Answering},
  author={Cai, Tianchi and Tan, Zhiwen and Song, Xierui and Sun, Tao and Jiang, Jiyan and Xu, Yunqi and Zhang, Yinger and Gu, Jinjie},
  booktitle={Proceedings of the 30th ACM SIGKDD Conference on Knowledge Discovery and Data Mining},
  pages={199--210},
  year={2024}
}

@article{zhang2025rag,
  title={RAG-Reward: Optimizing RAG with Reward Modeling and RLHF},
  author={Zhang, Hanning and Song, Juntong and Zhu, Juno and Wu, Yuanhao and Zhang, Tong and Niu, Cheng},
  journal={arXiv preprint arXiv:2501.13264},
  year={2025}
}

@inproceedings{gao2023enabling,
  title={Enabling Large Language Models to Generate Text with Citations},
  author={Gao, Tianyu and Yen, Howard and Yu, Jiatong and Chen, Danqi},
  booktitle={Proceedings of the 2023 Conference on Empirical Methods in Natural Language Processing},
  pages={6465--6488},
  year={2023}
}

@inproceedings{huang2024calibrating,
  title={Calibrating Long-form Generations From Large Language Models},
  author={Huang, Yukun and Liu, Yixin and Thirukovalluru, Raghuveer and Cohan, Arman and Dhingra, Bhuwan},
  booktitle={Findings of the Association for Computational Linguistics: EMNLP 2024},
  pages={13441--13460},
  year={2024}
}

@inproceedings{wang2022modeling,
  title={Modeling Exemplification in Long-form Question Answering via Retrieval},
  author={Wang, Shufan and Xu, Fangyuan and Thompson, Laure and Choi, Eunsol and Iyyer, Mohit},
  booktitle={Proceedings of the 2022 Conference of the North American Chapter of the Association for Computational Linguistics: Human Language Technologies},
  pages={2079--2092},
  year={2022}
}

@inproceedings{han2024rag,
  title={RAG-QA Arena: Evaluating Domain Robustness for Long-form Retrieval Augmented Question Answering},
  author={Han, Rujun and Zhang, Yuhao and Qi, Peng and Xu, Yumo and Wang, Jenyuan and Liu, Lan and Wang, William Yang and Min, Bonan and Castelli, Vittorio},
  booktitle={Proceedings of the 2024 Conference on Empirical Methods in Natural Language Processing},
  pages={4354--4374},
  year={2024}
}

\clearpage

\appendix

\section{Implementation Details}
\label{app:implementation}

To improve reproducibility, we provide the key training configuration of RioRAG in this appendix, and include the full prompt templates used in the ``{Extract}'' and ``{MergeCluster}'' stages of our reward construction pipeline.

\subsection{Training Hyperparameters}
\label{app:hyperparameters}

Following previous approaches~\cite{li2025viper,DBLP:conf/nips/ZhanLSG24}, We use a maximum prompt length of 8192 and a maximum response length of 4096. The training batch size is set to 64 with dynamic batch sizing enabled, and the maximum PPO token length per GPU is 12288. For optimization, we apply gradient clipping with a threshold of 1.0, use a clip ratio of 0.2, and set the entropy coefficient to 0.001. We enable KL regularization with \texttt{use\_kl\_loss=True}, using \texttt{low\_var\_kl} as the KL loss type and a KL coefficient of 0.04. PPO is run for one epoch without shuffling. We set the Ulysses sequence parallel size to 2. The learning rate is $1\times10^{-6}$, with a warmup ratio of 0.1 and cosine warmup scheduling. We will also release the full configuration file in our codebase. In the final version, we additionally report the hardware setup and random seed configuration explicitly.

\subsection{Details of Prompts}
\label{app:prompts}

In this appendix, we provide the exact prompt templates used in the ``{Extract}'' and ``{MergeCluster}'' stages of our reward construction pipeline. These prompts are used to convert retrieved web documents into concise factual nuggets and then merge them into a compact checklist for subsequent verification.

\paragraph{Extract Prompt.}
The ``{Extract}'' stage processes each retrieved web document independently. Given a user query and a retrieved web content, it outputs several highly relevant key points grounded only in the given content.

\paratitle{MergeCluster Prompt.}
After applying ``{Extract}'' to all retrieved documents, we obtain a list of candidate key points. The ``{MergeCluster}'' stage filters irrelevant items, removes exact duplicates, preserves complementary points with different focuses, and merges them into a final checklist.

\paratitle{Usage in Reward Construction.}
The output of ``Extract'' is a document-level set of candidate nuggets, while the output of ``MergeCluster'' is a query-level checklist used for informativeness assessment. This decomposition reduces the effective verification context length and provides a more compact and structured basis for reward computation.

\begin{tcolorbox}[colback=white,colframe=black,title=Extract Prompt]
You are given a user query and a retrieved web content.\\

Your task:\\
- Output several highly relevant key points from the web content.\\
- Each key point must be one concise sentence and separated by a newline.\\
- The key points must be wrapped inside \textbackslash boxed\{\\
Key Point 1\\
Key Point 2\\
...\\
\}.\\
- Only use the given web content.
- If no relevant information is found, output \textbackslash boxed\{No relevant information\}.\\
- Do not create or assume content not present in the web content.\\
\\
Input:\\
- Query: \{query\}\\
- Web Content:\\
\{web\_content\}\\
\\
- Query: \{query\}
\end{tcolorbox}

\section{Details on Datasets}
\label{sec:appendix_datasets}

In this section, we provide detailed descriptions of the two comprehensive benchmarks used in our experiments: {LongFact}~\cite{weilong} and {RAGChecker}~\cite{ru2024ragchecker}. These datasets are designed to evaluate long-form retrieval-augmented generation (RAG) systems across diverse topics and multiple dimensions of factual quality. Their complementary nature enables robust assessment of both factual coverage and fine-grained answer quality in open-domain settings.

\paratitle{LongFact.} 
LongFact is a manually curated benchmark focused on evaluating long-form factuality. It contains a diverse set of fact-seeking questions, where each gold answer synthesizes multiple atomic facts drawn from various evidence sources. The dataset is notable for its broad coverage across 38 fine-grained domains, which are grouped into the following 8 broader categories to support structured evaluation:

\begin{tcolorbox}[colback=white,colframe=black,title=MergeCluster Prompt]
You are given a user question and a list of candidate key points.\\
\\
Your task:\\
- Keep only the key points that are highly relevant to the question.\\
- Merge exact duplicates; if two points have slightly different focuses, keep both.\\
- Each item = one single idea, in one concise sentence.\\
- If there are conflicting views, use wording like: ``Some studies suggest [...], others indicate [...].''\\
- Output format: \textbackslash boxed\{\\
~~~~key point 1 \\
~~~~key point 2 \\
~~~~key point 3 \\
~~~~...\\
\}\\
- Only output inside \textbackslash boxed\{\}.\\
\\
Output Format:\\
\textbackslash boxed\{ \\
~~~~Final Key Point 1 \\
~~~~Final Key Point 2 \\
~~~~... \\
\}\\
\\
Input:\\
- Query: \{query\}\\
- Key Points:
\{key\_points\}\\
\\
- Query: \{query\}
\end{tcolorbox}

\begin{itemize}
    \item \emph{Science \& Nature}: physics, chemistry, biology, astronomy, virology, prehistory
    \item \emph{Technology \& Computing}: computer science, computer security, machine learning, electrical engineering, mathematics
    \item \emph{Medicine \& Psychology}: medicine, clinical knowledge, psychology, psychology checkpoint
    \item \emph{Law \& Politics}: international law, immigration law, U.S. foreign policy, jurisprudence
    \item \emph{Social Sciences \& Culture}: sociology, geography, world religions, moral disputes, philosophy
    \item \emph{History \& Events}: history, 20th-century events, global facts, economics
    \item \emph{Business \& Communication}: business ethics, accounting, marketing, management, public relations
    \item \emph{Entertainment \& Lifestyle}: movies, music, gaming, celebrities, architecture, sports
\end{itemize}

Each example in LongFact is annotated with atomic information units, enabling precise measurement of factual recall and information density. This makes it especially well-suited for evaluating long-form answers that integrate knowledge from multiple sources.

\paratitle{RAGChecker.} 
RAGChecker is a comprehensive benchmark designed to evaluate long-form Retrieval-Augmented Generation (RAG) systems across diverse domains. It repurposes examples from 10 public datasets, encompassing a total of 4,162 questions. For the 4 subsets we used, we briefly describe their characteristics below:

\begin{itemize}
    \item \emph{ClapNQ}~\cite{rosenthal2025clapnq}: Derived from Natural Questions (NQ), ClapNQ includes long-form answers with grounded gold passages from Wikipedia, focusing on generating cohesive long-form answers from non-contiguous text segments.
    \item \emph{NovelQA}~\cite{wang2024novelqa}: NovelQA is a benchmark designed to evaluate large language models on deep narrative understanding through complex questions based on English novels.
    \item \emph{FiQA}~\cite{maia201818}: A financial question answering dataset comprising 500 QA pairs, where short answers are extended to long-form using GPT-4, filtered to remove hallucinations.
    \item \emph{KIWI}~\cite{xu2024kiwi}: A dataset of knowledge-intensive writing instructions for answering research questions, comprising 71 QA pairs with long-form answers validated for quality.
\end{itemize}

\section{Details on Evaluation Metrics}
\label{sec:appendix_metrics}

We adopt two categories of metrics to comprehensively evaluate factual quality and retrieval grounding.

\paratitle{Standard LFQA Metrics.}
Following prior work~\cite{fan2019eli5}, we use \textit{Fact Recall} (FR) to measure factual completeness---the ratio of atomic facts in the generated response to those in the reference answer---and \textit{Information Density} (ID), defined as the ratio of atomic facts to total response length, reflecting conciseness and informativeness.

\paratitle{RAGChecker Metrics.}
The {RAGChecker} benchmark~\cite{ru2024ragchecker} introduces a suite of advanced metrics:
\textit{Faithfulness} (proportion of correct facts supported by retrieved pages),
\textit{Relevant Noise Sensitivity} (ratio of incorrect facts appearing in retrieved content),
\textit{Irrelevant Noise Sensitivity} (share of correct facts that are irrelevant to retrieval),
\textit{Hallucination} (incorrect facts unsupported by any retrieved page),
\textit{Self-Knowledge} (correct facts absent from all retrieved content), and
\textit{Context Utilization} (fraction of ground-truth facts covered by retrieval).
Together, these metrics provide a multi-dimensional evaluation of factual grounding, reliability, and retrieval efficiency.

% This is an appendix.

\end{document}